\documentclass[lettersize,journal]{IEEEtran}

\ifCLASSINFOpdf
\else
\usepackage[dvips]{graphicx}
\fi
\usepackage{url}

\usepackage{tabularx}
\usepackage{adjustbox}
\usepackage{amsfonts}       
\usepackage{amsmath}
\usepackage{amssymb}
\usepackage{amsthm}
\usepackage{algorithmic}
\usepackage{algorithm}
\usepackage{array}
\usepackage{arydshln}
\usepackage{bbm}
\usepackage{bm}
\usepackage{booktabs}       
\usepackage[font=small]{caption}
\usepackage{cite}
\usepackage{color}
\usepackage{colortbl}
\usepackage{diagbox}
\usepackage{float}
\usepackage{grffile} 
\usepackage{graphicx}
\usepackage{graphics}
\usepackage{makecell}
\usepackage{multirow}
\usepackage[subrefformat=parens,labelformat=parens]{subfig}
\usepackage{stfloats}
\usepackage{textcomp}
\usepackage{verbatim}
\usepackage{url}            
\usepackage{xcolor}

\ExplSyntaxOn
\NewDocumentCommand{\longdash}{ O{2} }
{
	--\prg_replicate:nn { #1 - 1 } { \negthinspace -- }
}
\ExplSyntaxOff

\usepackage{hyperref}       

\theoremstyle{plain}

\theoremstyle{definition}

\theoremstyle{remark}

\begin{document}
		
	\title{Regression-Oriented Knowledge Distillation for Lightweight Ship Orientation Angle Prediction with Optical Remote Sensing Images}
	
	\author{Zhan Shi, Xin Ding, Peng Ding, Chun Yang, Ru Huang and Xiaoxuan Song
		\thanks{Manuscript received \today; revised \today.}
		\thanks{Zhan Shi, Peng Ding, Chun Yang, Ru Huang, and Xiaoxuan Song are with Nanjing Research Institute of Electronic Engineering, Nanjing 210000, China (e-mail: z\_shi2006@163.com, dingpeng14@mails.ucas.ac.cn, yangguang326@126.com, rdray@126.com, 284214209@qq.com).}
		\thanks{Xin Ding is with the Department of Artificial Intelligence, Nanjing University of Information Science \& Technology, Nanjing 210044, China (e-mail: 003763@nuist.edu.cn) \textit{(Corresponding author: Xin Ding)}.}
		\thanks{Zhan Shi and Xin Ding contribute equally to this paper.}}
	
	\markboth{Preprint. Under Review.}
	{Shell \MakeLowercase{\textit{et al.}}: Bare Demo of IEEEtran.cls for IEEE Journals}
	\maketitle
	
	\begin{abstract}
		
		Ship orientation angle prediction (SOAP) with optical remote sensing images is an important image processing task, which often relies on deep convolutional neural networks (CNNs) to make accurate predictions. This paper proposes a novel framework to reduce the model sizes and computational costs of SOAP models without harming prediction accuracy. First, a new SOAP model called Mobile-SOAP is designed based on MobileNetV2, achieving state-of-the-art prediction accuracy. Four tiny SOAP models are also created by replacing the convolutional blocks in Mobile-SOAP with four small-scale networks, respectively. Then, to transfer knowledge from Mobile-SOAP to four lightweight models, we propose a novel knowledge distillation (KD) framework termed SOAP-KD  consisting of a novel feature-based guidance loss and an optimized synthetic samples-based knowledge transfer mechanism. Lastly, extensive experiments on the FGSC-23 dataset confirm the superiority of Mobile-SOAP over existing models and also demonstrate the effectiveness of SOAP-KD in improving the prediction performance of four specially designed tiny models. Notably, by using SOAP-KD, the test mean absolute error of the ShuffleNetV2$\times$1.0-based model is only 8\% higher than that of Mobile-SOAP, but its number of parameters and multiply–accumulate operations (MACs) are respectively 61.6\% and 60.8\% less. Our codes can be found at \url{https://github.com/UBCDingXin/SOAP-KD}.
		
	\end{abstract}
	

	\IEEEpeerreviewmaketitle

	\section{Introduction}

	\IEEEPARstart{S}{hip} \textit{orientation angle prediction} (SOAP) is an important image processing task in \textit{remote sensing} (RS), playing an essential role in ship traffic monitoring, maritime surveillance, and naval warfare. A suitable SOAP method can substantially benefit ship detection, especially when adjacent ships dock closely, by helping build minimum enclosing bounding boxes \cite{ma2019ship, niu2020learning, niu2022efficient, shi2014,hua2019}. Furthermore, it can also be used to forecast the direction of a target ship's navigation \cite{wang2018simultaneous}, resulting in accurate ship tracking. Additionally, the SOAP task is also able to improve the accuracy of fine-grained ship classification \cite{zhang2020new}. 
	
	With the rapid development of deep learning, most SOAP methods are developed based on deep neural networks and achieve high precision on optical RS images. Instead of performing SOAP alone, most of these methods conduct the orientation prediction along with ship detection or classification. For example, Yang et al.~\cite{yang2018automatic, yang2018position} relied on the rotated bounding boxes from a ship detection task to calculate ship orientation angles. Ma et al.~\cite{ma2019ship} proposed to convert SOAP into a classification problem by binning orientation angles into $K$ classes and then use an extra branch in a U-shape ship detection network to perform such classification. Zhang et al.~\cite{zhang2020new} recently developed an \textit{attribute-guided multilevel enhanced feature representation network} (AMEFRN), where the SOAP result is taken as auxiliary information for better ship classification. Niu et al.~\cite{niu2020learning} proposed an annotation-free algorithm based on the Hough transform and a pre-trained classification \textit{convolutional neural network} (CNN), which performs SOAP in an unsupervised manner. However, all the above methods rely heavily on deep neural networks (e.g., VGG16~\cite{simonyan2014very}) as the backbone for accurate predictions, where large model size and high computational cost prevent them from being deployed on devices with limited computational resources, e.g., unmanned aerial vehicles, airship, and satellite. 
	
	\textit{Knowledge distillation} (KD) is an effective technique for model compression and has been widely applied in optical RS image processing. For instance, in the scene classification tasks, Xu et al.~\cite{xu2022vision} defined a logits-based KD loss to transfer knowledge from a heavyweight vision transformer (i.e., teacher) to a lightweight CNN (i.e., student). Li et al.~\cite{li2022remote} performed KD in the same task by introducing two new network modules and the corresponding feature-based KD losses for knowledge transfer. For the object detection task, Yang et al.~\cite{yang2022statistical} introduced a new imitation mechanism to distill core information in the teacher's features and a regression distillation module encouraging the student model to mimic accurate detection results from the teacher model. Moreover, for the change detection task, Mahmoud et al.~\cite{mahmoud2021training} proposed a logits-based mechanism to transfer knowledge from a large Siamese teacher network to a tiny student network. Unfortunately, the above KD methods often rely on soft class labels in classification, bounding boxes in object detection, or specially designed network modules to transfer knowledge. Thus, they are either entirely inapplicable or have some modules invalid in SOAP.

	Ding et al.~\cite{ding2023distilling} recently proposed an effective KD method called cGAN-KD for the natural image regression with a scalar response variable. Instead of defining extra KD losses or modifying network architectures, cGAN-KD utilizes fake samples generated from \textit{continuous conditional generative adversarial networks} (CcGANs)~\cite{ ding2022continuous} to transfer knowledge and has been applied to face recognition and autonomous driving. cGAN-KD is also applicable to SOAP with optical RS images, where SOAP is formulated as a regression task with the orientation angle as the scalar response variable.
	
	Motivated by the above problems and enlightened by cGAN-KD, we propose a simple but effective KD method for SOAP termed \textbf{SOAP-KD}, making the deployment of precise and lightweight SOAP models on edge devices possible. Our contributions can be summarized as follows. First, we designed an accurate SOAP model termed \textbf{Mobile-SOAP} consisting of the MobileNetV2 convolutional blocks pre-trained on ImageNet and followed by three fully-connected layers. Second, we designed four tiny CNNs whose model sizes and computation costs are substantially smaller than existing SOAP models. Third, we proposed SOAP-KD to transfer knowledge from Mobile-SOAP to four tiny SOAP models by combining an optimized cGAN-KD framework with a feature-based KD loss. Lastly, we conduct extensive experiments on FGSC-23 to show that Mobile-SOAP outperforms existing SOAP models and SOAP-KD can effectively improve tiny SOAP models' precision by using Mobile-SOAP's knowledge.

%
%

	\section{Methodology}

	\begin{figure*}[!htbp]
		\centering
		\includegraphics[width=0.8\textwidth]{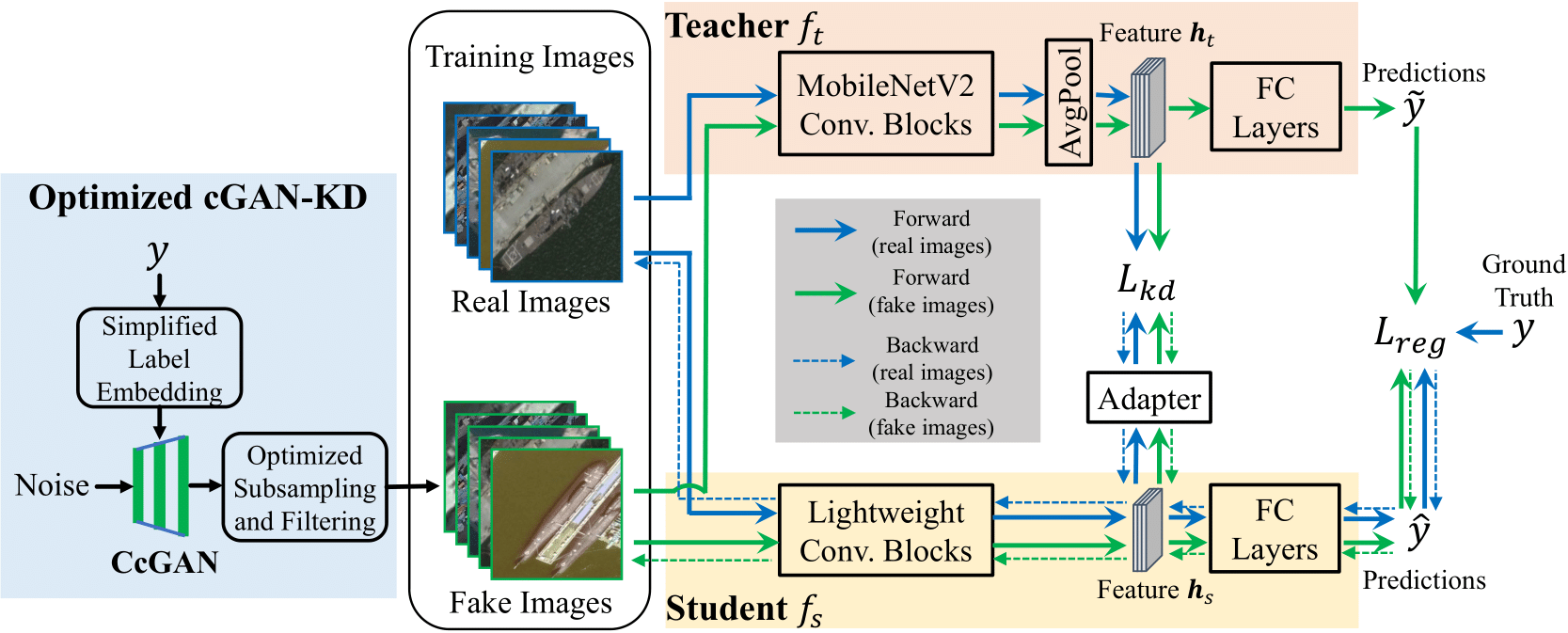}
		\caption{The overall framework of the proposed SOAP-KD method.}
		\label{fig:overall_workflow}
	\end{figure*}
	
	\subsection{Problem Formulation}
	
\begin{figure}[!htbp]
		\centering
		\subfloat[{\small An example optical remote sensing image $\bm{x}$ with orientation angle $y$.}]{\includegraphics[width=0.4\linewidth]{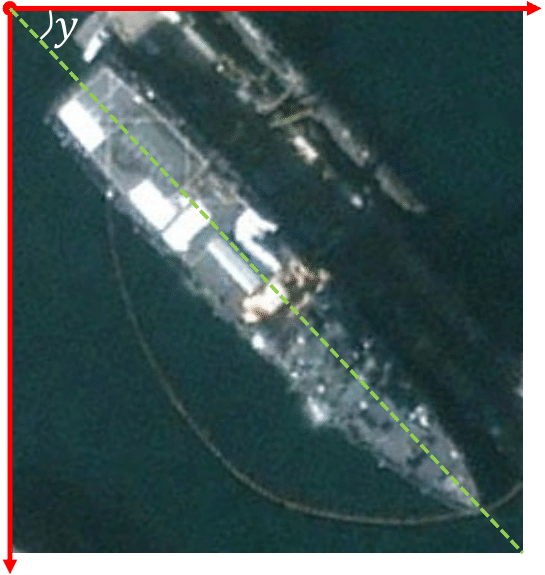}%
			\label{fig:img_example}}
		\hfil
		\subfloat[{\small The range of $y$ is $[0^\circ,180^\circ]$.}]{\includegraphics[width=0.5\linewidth]{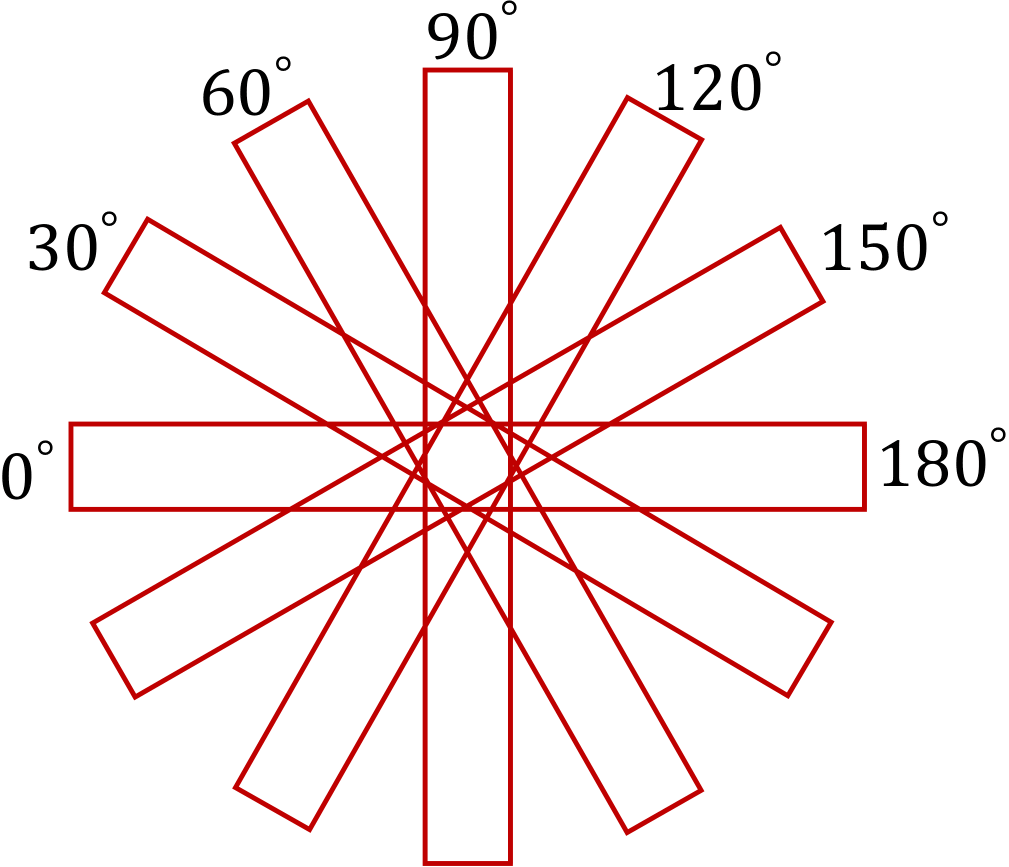}%
			\label{fig:angle_example}}
		\caption{An illustrative example of ship orientation angle prediction.}
		\label{fig:problem_formulation}
	\end{figure}

	A SOAP model $f(\cdot)$ aims to accurately predict the orientation angle $y$ of a ship based on its optical RS image $\bm{x}$ (see Fig.~\subref*{fig:img_example}), which can be formulated as image regression with a scalar response variable. The image $\bm{x}$ is an RGB image and is assumed at $224\times 224$ resolution in this paper. Since Niu et al.~\cite{niu2020learning} pointed out that it is challenging to differentiate the heads and tails of some ships from RS images, such as container ships, we ignore the distinction between the bow or stern in SOAP. In that case, $y$ ranges from $0^\circ$ to $180^\circ$.
	
	State-of-the-art SOAP models~\cite{ma2019ship, zhang2020new} rely on deep CNNs to make accurate predictions, resulting in large model sizes and high computational costs. This paper focuses on developing a lightweight and efficient SOAP model by transferring the ``dark knowledge" from a pre-trained heavyweight SOAP model (aka teacher) to a tiny regression CNN (aka student).

	\subsection{Overall Framework of SOAP-KD}
	
	The overall framework of the proposed SOAP-KD method can be summarized in Fig.~\ref{fig:overall_workflow}. It mainly consists of the following components: a pre-trained accurate teacher model $f_t$, a tiny student model $f_s$, an optimized cGAN-KD method, and a feature-based guidance mechanism. The last two components encourage the student to mimic the teacher's performance to improve the student's prediction accuracy.

	\subsection{Accurate Mobile-SOAP and Lightweight Students}
	\label{sec:teacher_and_students}
	
\begin{figure}[!htbp]
		\centering
		\includegraphics[width=0.9\linewidth]{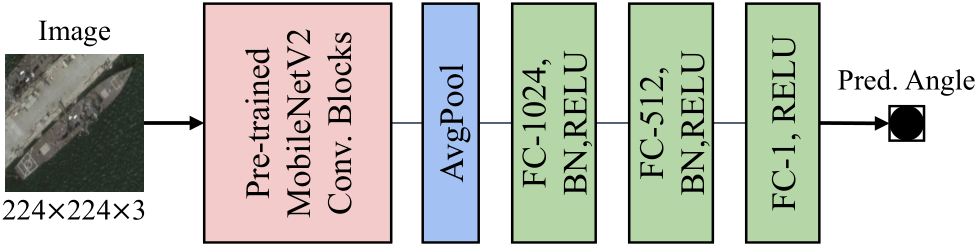}
		\caption{The architecture of Mobile-SOAP (the teacher in SOAP-KD).}
		\label{fig:teacher_architecture}
	\end{figure}

	In this section, we first propose a simple but accurate teacher model called Mobile-SOAP visualized in Fig.~\ref{fig:teacher_architecture} to replace ASD~\cite{ma2019ship} and AMEFRN~\cite{zhang2020new}. Mobile-SOAP starts with MobileNetV2's convolutional blocks pre-trained on ImageNet and ends with three fully-connected layers. Although with less requirement on computational resources, Mobile-SOAP is more precise than ASD and AMEFRN whose backbone networks are based on VGG16~\cite{simonyan2014very}. Furthermore, we propose four tiny SOAP models whose architectures are similar to that in Fig.~\ref{fig:teacher_architecture} but replace the MobileNetV2 blocks with the convolutional layers of ResNet8~\cite{he2016deep}, WRN16$\times$1~\cite{zagoruyko2016wide}, ShuffleNetV2$\times$0.5~\cite{ma2018shufflenet}, and ShuffleNetV2$\times$1.0~\cite{ma2018shufflenet}, respectively. Compared with ASD, AMEFRN, and Mobile-SOAP, these four tiny models' precisions are worse but need much less computational costs. The number of parameters and multiply–accumulate operations (MACs) of Mobile-SOAP and four tiny CNNs are shown in Table~\ref{tab:soap_results}.

	\subsection{Optimized cGAN-KD}
	
\begin{figure}[!htbp]
		\centering
		\includegraphics[width=1\linewidth]{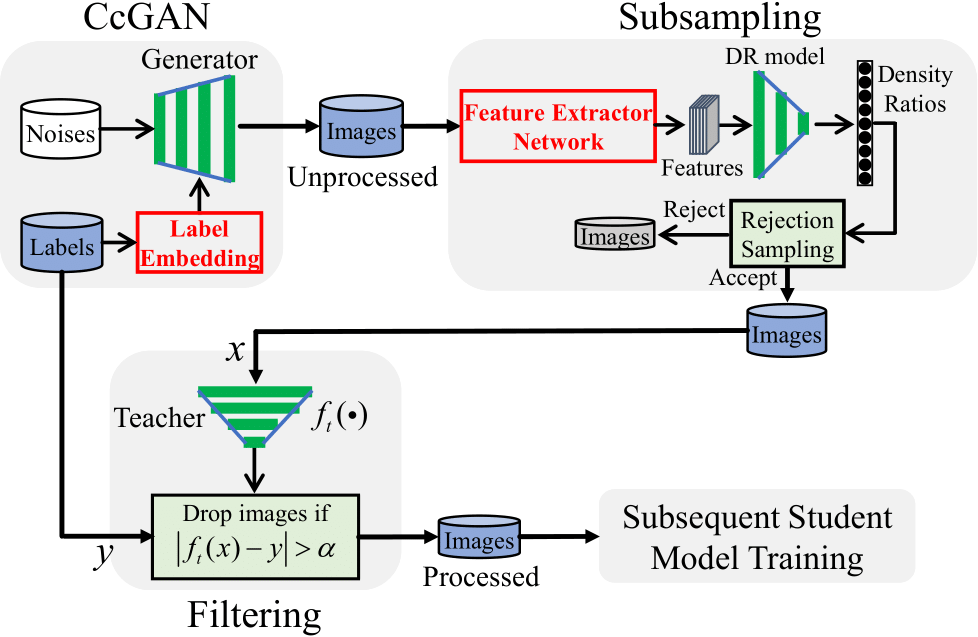}
		\caption{The workflow of cGAN-KD (adapted from~\cite{ding2023distilling}).}
		\label{fig:workflow_cGAN-KD}
	\end{figure}

\begin{figure}[!htbp]
		\centering
		\includegraphics[width=0.9\linewidth]{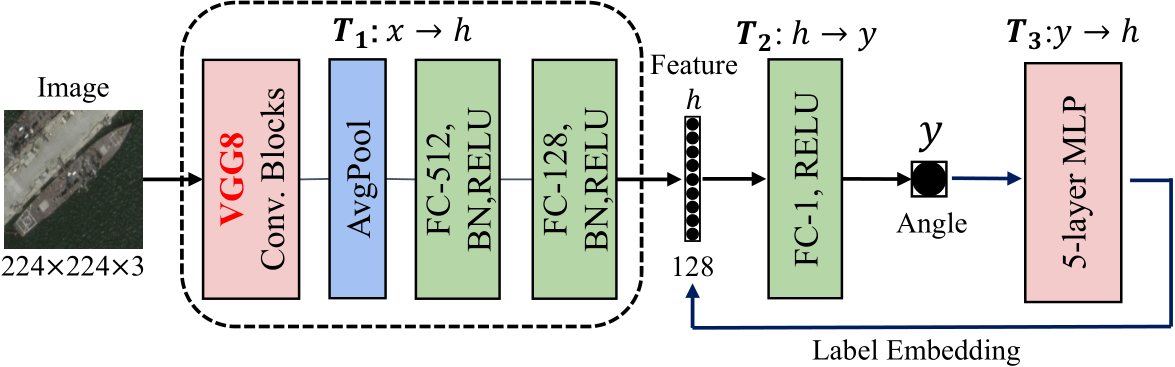}
		\caption{The simplified label embedding mechanism for CcGAN. To train the label embedding network $T_3$, we pre-train $T_1+T_2$ on the training set, where we adopt VGG8 instead of ResNet34 as the backbone for $T_1$ and reduce the training epochs from 200 to only 10.}
		\label{fig:simple_label_embed}
	\end{figure}

\begin{figure}[!htbp]
		\centering
		\includegraphics[width=1\linewidth]{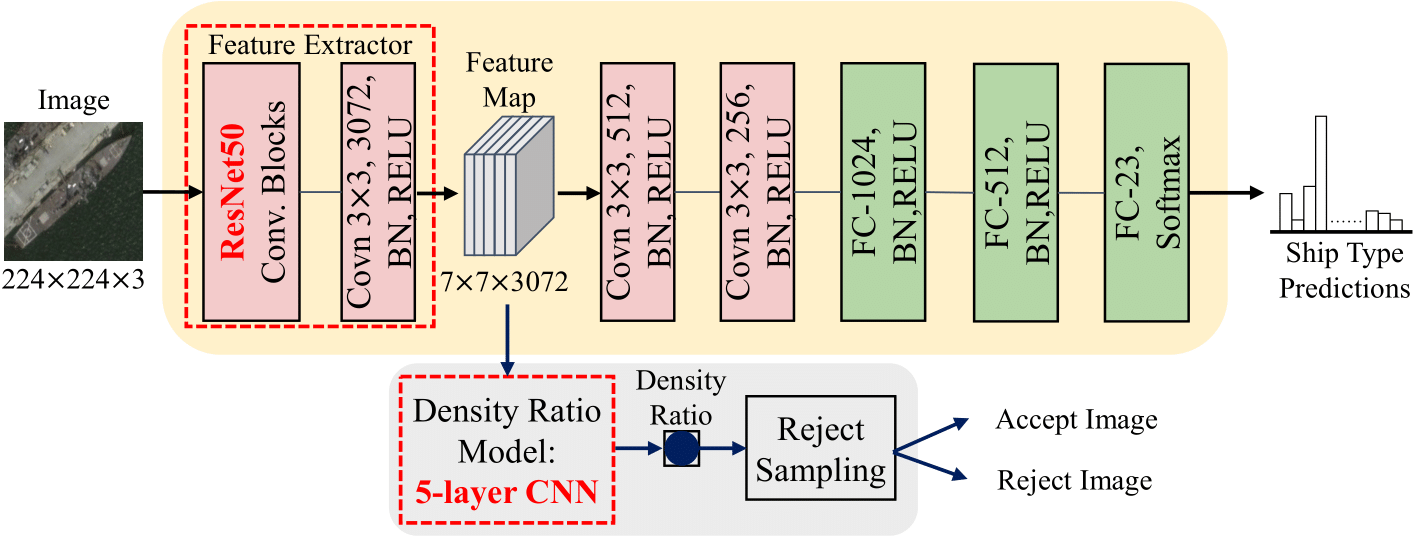}
		\caption{The pipeline of the optimized subsampling module in cGAN-KD, including a ResNet50-based feature extractor and a 5-layer CNN-based density ratio mode.}
		\label{fig:optimized_subsampling}
	\end{figure}

	To improve the performance of four tiny SOAP models proposed above, we introduce cGAN-KD~\cite{ding2023distilling} into SOAP, a regression-oriented KD designed for image regression with a scalar response variable. The pipeline of cGAN-KD is summarized in Fig.~\ref{fig:workflow_cGAN-KD} and can be split into three sequential steps, including CcGAN, Subsampling, and Filtering. Before applying cGAN-KD to SOAP, we optimize the label embedding mechanism of CcGAN and the subsampling module, which are visualized in Figs.~\ref{fig:simple_label_embed} and~\ref{fig:optimized_subsampling}, respectively. 
	
	To be specific, \cite{ding2023distilling} first trained a regression CNN (i.e., $\bm{T}_1+\bm{T}_2$ in Fig~\ref{fig:simple_label_embed}) to encode regression labels in the CcGAN training, where $\bm{T}_1$ consists of ResNet34's convolutional blocks and $\bm{T}_2$ defines a mapping from latent features $\bm{h}$ to regression labels $y$. With the fixed $\bm{T}_2$, \cite{ding2023distilling} then trained 5-layer multilayer perceptron (MLP) to map regression labels $y$ back to their latent representations $\bm{h}$, i.e., $\bm{T}_3$, by minimizing 
	\begin{equation*}
		\min_{\bm{T}_3}\mathbb{E}_{y\sim p(y)}\mathbb{E}_{\gamma\sim\mathcal{N}(0,0.02)}\left[ (\bm{T}_2(\bm{T}_3(y+\gamma)) - (y+\gamma))^2 \right],
	\end{equation*}
	so that $\bm{T}_3$, as the label embedding network, defines an one-to-one mapping from $y$ to $\bm{h}$. However, using ResNet34 in this mechanism is redundant since the accuracy of $\bm{T}_1+\bm{T}_2$ won't affect the label embedding performance. Therefore, we replace ResNet34 with VGG8 and reduce the training epochs from 200 to only 10 to simplify the training process.

	Furthermore, \cite{ding2023distilling} conducted reject sampling to select high-quality fake images generated from CcGAN based on conditional density ratio estimation (DRE)~\cite{ding2023efficient} (aka \textbf{subsampling}). As a preliminary step of subsampling, \cite{ding2023distilling} trained a sparse autoencoder (SAE) to extract high-level features from images since many regression datasets do not have categorical notations, and the features' dimension is designed to be consistent with input images. Then, a density ratio model with five fully-connected layers (aka MLP-5) is trained in the feature space to estimate the conditional density ratio of a given image. However, images in SOAP datasets are often annotated by ship types, and the SAE used by \cite{ding2023distilling} may suffer from overfitting even with regularization due to the ``equal dimension requirement". Meanwhile, the model size of the MLP-5-based DRE model is often too large when images are high-resolution. Fortunately, many works~\cite{zhang2020new,chen2022contrastive} show that ResNet50 is a good backbone for ship classification and \cite{ding2023efficient} reports that DRE benefits from precise classification CNNs-based feature extractors. Besides, the weight sharing mechanism in CNN can substantially reduce model size. Therefore, we propose replacing SAE with a ResNet50-based classifier for feature extraction and using a 5-layer CNN as the DRE model (Fig.~\ref{fig:optimized_subsampling}).

	The last module in cGAN-KD is Filtering, which is unchanged in SOAP-KD. Filtering uses the teacher model $f_t$ to predict the labels of fake images and drops those with predicted labels far from the conditioning labels in CcGAN. Filtering is performed based on a data-dependent threshold $\alpha$~\cite{ding2023distilling}, and the remaining fake images are used to train student models in a data augmentation manner.
		
	\subsection{Feature-based Guidance}
	
	Besides cGAN-KD, as shown in Fig.~\ref{fig:overall_workflow}, we also design a KD loss $L_{kd}$ to match the features extracted by the convolutional blocks of the teacher and student as follows:	
	\begin{equation}
		L_{kd} = \sum_{i=1}^{N^r}\|\bm{h}^{r}_{t,i}-\phi(\bm{h}^{r}_{s,i})\|_2^2 + \sum_{i=1}^{N^g}\|\bm{h}^{g}_{t,i}-\phi(\bm{h}^{g}_{s,i})\|_2^2. 
		\label{eq:L_kd}
	\end{equation}
	In Eq.~\eqref{eq:L_kd}, $\bm{h}^{r}_{t,i}$, $\bm{h}^{r}_{t,i}$, $\bm{h}^{g}_{t,i}$, and $\bm{h}^{g}_{s,i}$ are extracted features, $r$ and $g$ specify the real and fake images, $N^r$ and $N^g$ denote the sample sizes of real and fake images. To eliminate the dimension difference between the features extracted from teacher and student, inspired by \cite{romero2014fitnets}, we propose an adapter network with one convolutional layer to adjust the student's features. $L_{kd}$ functions as a regularizer, encouraging the student to behave similarly to the teacher.

	\subsection{Training Loss}
	
	The regression loss for SOAP is defined as follows:
	\begin{equation}
		L_{reg} = \sum_{i=1}^{N^r}(f_s(\bm{x}^r_i)-y^r_i)^2 + \sum_{i=1}^{N^g}(f_s(\bm{x}^g_i)-f_t(\bm{x}^g_i))^2,
		\label{eq:L_reg}
	\end{equation}
	where $\bm{x}^r_i$ and $\bm{x}^g_i$ are real and fake images, $y^r_i$ is the ground truth angle for $\bm{x}^r_i$, the teacher $f_t$ performs pseudo labeling on fake images, and $f_s$ stands for the student. The final training loss for SOAP-KD is 
	\begin{equation}
		L = L_{reg} + \lambda L_{kd}.
		\label{eq:obj}
	\end{equation}
	The optimal hyper-parameter $\lambda$ can be selected by grid search (e.g., $[0.1, 1,10,100]$) on a validation set, where fake images are not included in training to reduce computational cost.

	\section{Experimental evaluation}
	
	We empirically demonstrate the effectiveness of SOAP-KD in model compression on the FGSC-23 dataset. 
	
	\subsection{FGSC-23 Dataset}
	
	FGSC-23 dataset \cite{yao2021fgsc, chen2022contrastive} is a popular remote sensing dataset for fine-grained ship classification and ship orientation angle prediction. It consists of high-resolution optical remote sensing images for 23 types of ships. By default, FGSC-23 is split into a training set with 6512 images and a test set with 1650 images. To select the optimal $\lambda$ in Eq.~\eqref{eq:obj}, we further randomly split the training set into a sub-training set and a validation set with a ratio of 8:2. Following \cite{zhang2020new}, we add blank pixels to non-square images to make them square, and then resize all images to $224\times 224$.

	\subsection{Experimental Setups}
	
	The proposed method is compared with two state-of-the-art SOAP models (ASD~\cite{ma2019ship} and AMEFRN~\cite{zhang2020new}) in terms of model size (\# Params), computational cost (MACs), and the test mean absolute error (MAE). When implementing ASD, we bin angles into 60 disjoint intervals (i.e., 60 classes).  Some works, such as \cite{wang2018simultaneous, niu2022efficient}, are designed for Synthetic aperture radar (SAR) images and not general enough for optical RS images, so they are not included in the comparison. Furthermore, we also compare SOAP-KD with other KD methods, including FitNet~\cite{romero2014fitnets}, RKD~\cite{zhao2020distilling}, and DKD~\cite{li2022remote}. Note that FitNet, RKD, and DKD are all initially designed for classification, so some of their modules or loss functions are invalid in SOAP. Therefore, we modify these KD methods to fit our experimental setting. Moreover, two ablation studies are also performed to analyze the main components of SOAP-KD. The first ablation study investigates the effects of different teacher's backbones, and the second one is used to test the effectiveness of $L_{kd}$ and the optimized cGAN-KD method. When training SOAP models, the epochs and batch size are set to 200 and 128, respectively, the initial learning rate is 0.01, and the learning rate decays at 80-th and 150-th epoch, respectively.  Please see Appendix and our codes for more details about the experimental setups.

	\subsection{Experimental Results}
	
\begin{figure}[!htbp]
		\centering
		\includegraphics[width=0.75\linewidth]{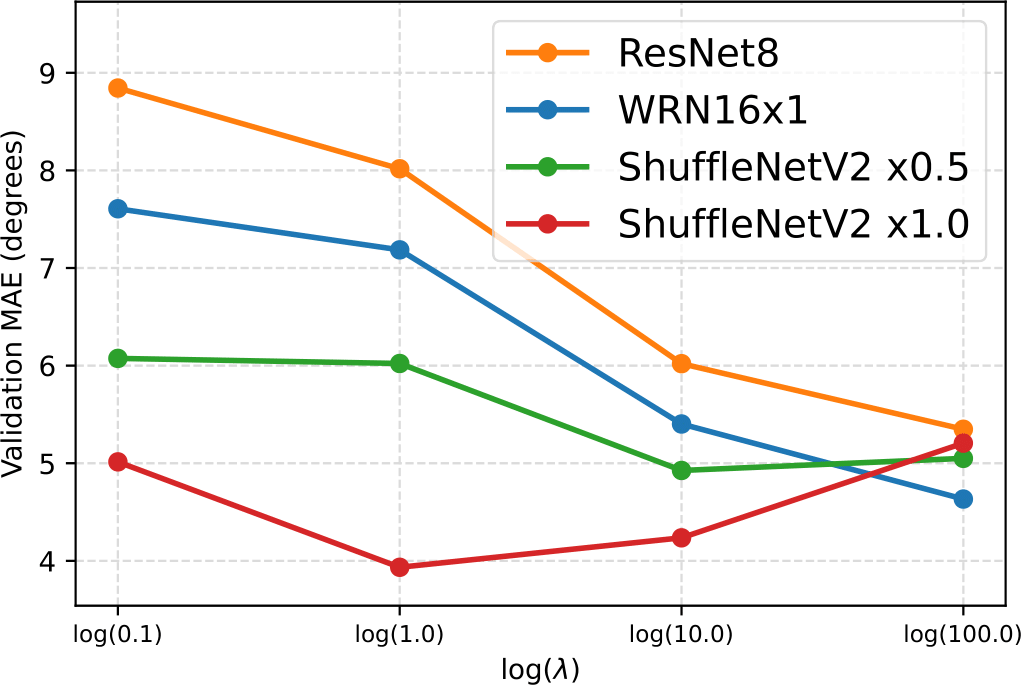}
		\caption{Selecting the optimal $\lambda$ when implementing SOAP-KD with different students on FGSC-23 by minimizing their validation MAEs.}
		\label{fig:fgsc23_validation_results_lambda}
	\end{figure}

	Table~\ref{tab:soap_results} shows that the proposed Mobile-SOAP outperforms ASD and AMEFRN by a large margin for all three evaluation metrics. With SOAP-KD, the test MAEs of ShuffleNetV2$\times$1.0 and WRN16$\times$1 are comparable to that of Mobile-SOAP, but these two tiny students require much less computational costs. \textbf{Notably, the test MAE of ShuffleNetV2$\times$1.0 is only 8\% higher than that of Mobile-SOAP, but its \# Params and MACs are respectively 61.6\% and 60.8\% less than those of Mobile-SOAP.} The validation results of students under different $\lambda$'s are also shown in Fig.~\ref{fig:fgsc23_validation_results_lambda}.
	
	Table~\ref{tab:kd_results} shows the test results of different KD methods, where SOAP-KD outperforms all three KD methods under all teacher-student combinations. Notably, some KD methods are worse than NOKD under some teacher-student combinations. 
	
\begin{figure}[!htbp]
		\centering
		\includegraphics[width=0.75\linewidth]{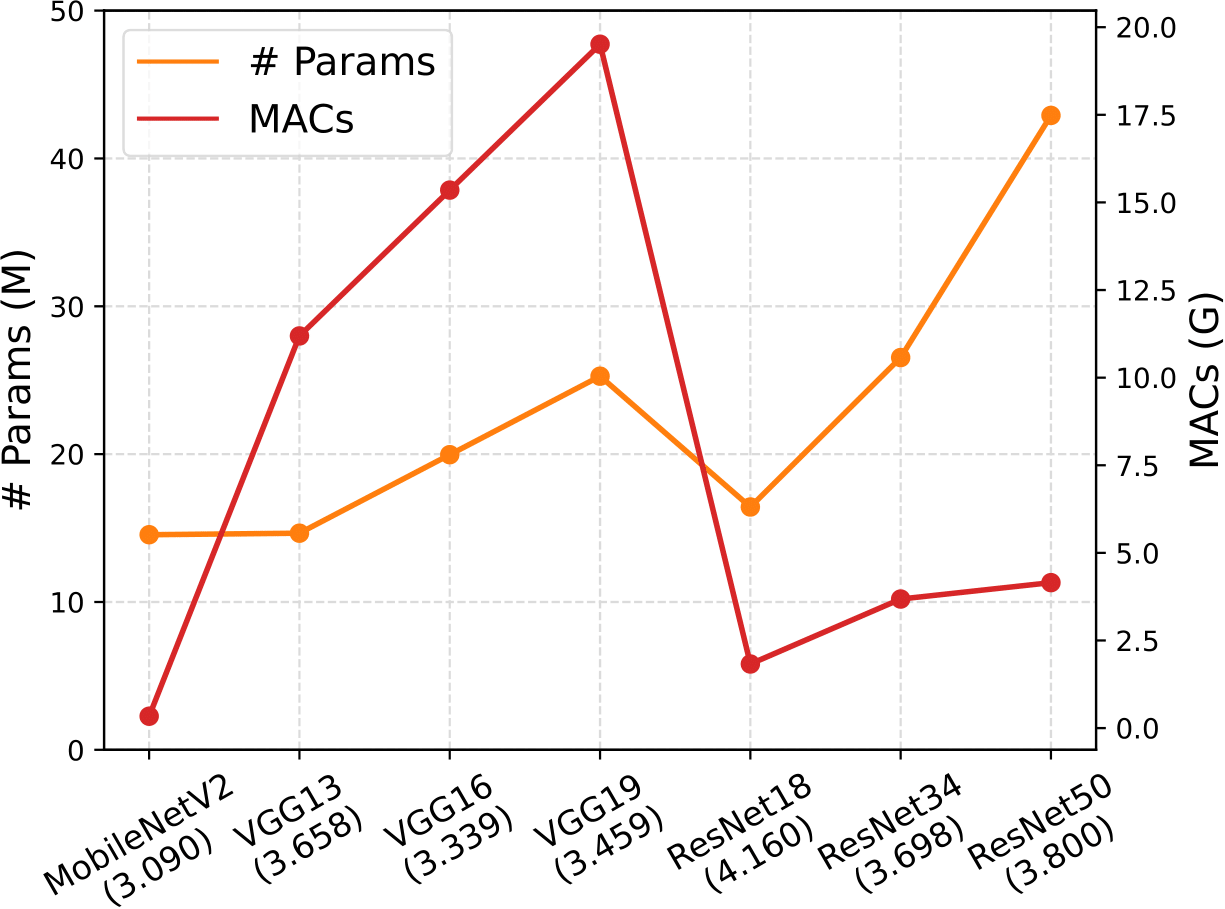}
		\caption{Ablation Study I: The \# Params and MACs of different teacher's backbones with their test MAEs in the paratheses.}
		\label{fig:fgsc23_test_results_teacher}
	\end{figure}

	The empirical results of two ablation studies are shown, respectively, in Fig.~\ref{fig:fgsc23_test_results_teacher} and Table~\ref{tab:ablation_diff_components}. Although MobileNetV2, as the teacher's backbone, has the fewest parameters and MACs, it also outperforms other backbone networks in test MAE, implying MobileNetV2 is a good choice for building the teacher in SOAP-KD. Table~\ref{tab:ablation_diff_components} proves that the combination of $L_{kd}$ and the optimized cGAN-KD leads the best KD result. 
	

	\begin{table}[!htbp]
		\centering
		\caption{The comparison of different SOAP models based on their model sizes, computational costs, and test MAEs on FGSC-23. Notably, the test MAE of ShuffleNetV2$\times$1.0 is only 8\% higher than that of Mobile-SOAP, but its \# Params and MACs are respectively 61.6\% and 60.8\% less than Mobile-SOAP.}
		\begin{adjustbox}{width=0.9\linewidth}
		\begin{tabular}{lccc}
			\toprule
			\textbf{Models} & \begin{tabular}[l]{@{}c@{}} \textbf{\# Params} $\downarrow$\\ ($\times 10^6$)\end{tabular} & \begin{tabular}[l]{@{}c@{}} \textbf{MACs} $\downarrow$ \\ ($\times 10^9$)\end{tabular} & \begin{tabular}[l]{@{}c@{}} \textbf{Test MAE} $\downarrow$\\ (degrees)\end{tabular} \\
			\midrule
			\textbf{Existing Models} & & & \\
			ASD (60 classes) (2019) & 40.964  & 15.373  & 4.167  \\
			AMEFRN (2020) & 40.934  & 15.373  & 3.784  \\
			
			\midrule
			\textbf{Proposed Teacher} & & & \\
			Mobile-SOAP & 14.550  & 0.339  & 3.090 \\
			
			\midrule
			\textbf{Proposed Students} & & & \\
			ResNet8 w/ SOAP-KD & 1.197 & 0.104  & 4.192  \\
			WRN16$\times$1 w/ SOAP-KD  & 0.390 & 0.182  & 3.514  \\
			ShuffleNetV2$\times$0.5 w/ SOAP-KD & 2.442  & 0.036 & 3.718 \\
		    ShuffleNetV2$\times$1.0 w/ SOAP-KD & 5.582  & 0.133 & 3.339 \\
			\bottomrule
		\end{tabular}%
		\end{adjustbox}
		\label{tab:soap_results}%
	\end{table}%

	\begin{table}[!htbp]
		\centering
		\caption{The test MAE comparison of different KD methods on FGSC-23 with four lightweight students and a fixed teacher (teacher is Mobile-SOAP whose test MAE is $3.09$ degrees). NOKD means no KD method is applied.}
		\begin{adjustbox}{width=1\linewidth}
		\begin{tabular}{lcccc}
			\toprule
			\diagbox{\textbf{Methods}}{\textbf{Students}} & \textbf{ResNet8} & \textbf{WRN16$\times$1} & \textbf{ShuffleNetV2$\times$0.5} & \textbf{ShuffleNetV2$\times$1.0} \\
			\midrule
			NOKD  & 8.941   & 8.017  & 5.222  & 5.119  \\
			\midrule
			FitNet (2015) & 4.977   & 4.586  & 5.524  & 5.766  \\
			RKD (2019) & 6.881   & 5.586  & 5.418  & 4.864  \\
			DKD (2022) & 6.175   & 6.410  & 5.520  & 4.639  \\
			\midrule
			SOAP-KD (ours) & \textbf{4.192} &  \textbf{3.514} & \textbf{3.718} & \textbf{3.339} \\
			\bottomrule
		\end{tabular}%
		\end{adjustbox}
		\label{tab:kd_results}%
	\end{table}%

	
	\begin{table}[!htbp]
		\centering
		\caption{Ablation Study II: The influence of the feature-based guidance loss and the cGAN-KD method.}
		\begin{adjustbox}{width=1\linewidth}
			\begin{tabular}{lcccc}
				\toprule
				\diagbox{\textbf{Methods}}{\textbf{Students}} & \textbf{ResNet8} & \textbf{WRN16$\times$1} & \textbf{ShuffleNetV2$\times$0.5} & \textbf{ShuffleNetV2$\times$1.0} \\
				\midrule
				NOKD  & 8.941   & 8.017  & 5.222  & 5.119  \\
				\midrule
				$L_{kd}$ & 4.977  & 4.586  & 4.105  & 4.023  \\
				$L_{kd}$ + cGAN-KD & \textbf{4.192} & \textbf{3.514} & \textbf{3.718} & \textbf{3.339} \\
				\bottomrule
			\end{tabular}%
		\end{adjustbox}
		\label{tab:ablation_diff_components}%
	\end{table}%

	\section{Conclusion}
	
	In this paper, we first designed a new regression CNN called Mobile-SOAP, achieving state-of-the-art prediction performance in the SOAP task. Meanwhile, we also designed four lightweight SOAP models as students. Then, we proposed a novel KD framework termed SOAP-KD to transfer knowledge from the accurate Mobile-SOAP to four students. Extensive experiments on FGSC-23 demonstrate that Mobile-SOAP outperforms newest SOAP models with a large margin, based on which SOAP-KD can substantially enhance the prediction accuracy of four tiny models. These efficient and accurate SOAP models may effectively benefit downstream ship detection or tracking tasks.

	\bibliographystyle{IEEEtran}
	\bibliography{reference}

\end{document}